\documentclass[10pt,twocolumn,letterpaper]{article}

\usepackage{iccv}
\usepackage{times}
\usepackage{epsfig}
\usepackage{graphicx}
\usepackage{amsmath}
\usepackage{amssymb}
\usepackage{booktabs}
\usepackage{multirow}
\usepackage{subfigure}
\newcommand{\dlt}[1]{}
\newcommand{\vb}[1]{\mathbf{#1}}
\newcommand{\ppf}{\mathcal{F}} 

\usepackage[breaklinks=true,bookmarks=false]{hyperref}

\iccvfinalcopy 

\ificcvfinal\fi

\begin{document}

\title{DPF-Net: Combining Explicit Shape Priors in Deformable Primitive Field \\ for Unsupervised Structural Reconstruction of 3D Objects}

\author{ 
Qingyao Shuai, Chi Zhang, Kaizhi Yang, Xuejin Chen\thanks{Corresponding author}\\ 
National Engineering Laboratory for Brain-inspired Intelligence Technology and Application
\\
University of Science and Technology of China\\
 {\tt\small \{qyshuai21, chih, ykz0923\}@mail.ustc.edu.cn, xjchen99@ustc.edu.cn }
}

\maketitle
\ificcvfinal\thispagestyle{empty}\fi

\begin{abstract}

Unsupervised methods for reconstructing structures face significant challenges in capturing the geometric details with consistent structures among diverse shapes of the same category.
To address this issue, we present a novel unsupervised structural reconstruction method, named DPF-Net, based on a new Deformable Primitive Field (DPF) representation, which allows for high-quality shape reconstruction using parameterized geometric primitives.  
We design a two-stage shape reconstruction pipeline which consists of a primitive generation module and a primitive deformation module to approximate the target shape of each part progressively.
The primitive generation module estimates the explicit orientation, position, and size parameters of parameterized geometric primitives, while the primitive deformation module predicts a dense deformation field based on a parameterized primitive field to recover shape details. 
The strong shape prior encoded in parameterized geometric primitives enables our DPF-Net to extract high-level structures and recover fine-grained shape details consistently. 
The experimental results on three categories of objects in diverse shapes demonstrate the effectiveness and generalization ability of our DPF-Net on structural reconstruction and shape segmentation.
 
\end{abstract}

\section{Introduction}
\label{Sec.intro}

\dlt{
\begin{figure}
  \centering
  \includegraphics[width=0.5\textwidth]{figures/intro.pdf}
  \caption{The structural representation for 3D objects. Previous primitive methods~\cite{VP-CVPR-2017,HA-Siggraph-2019,CA-Siggraph-2021} have difficulty to modeling detailed geometry of objects, the implicit function methods~\cite{ BAENet-ICCV-2019, RIM-CVPR-2022} suffered the lack of explainable structure extraction. We propose DPF, a novel  structural representation for 3D objects which can represent both concise and explainable structure and detailed geometry.}
  \label{fig.intro} 
  \vspace{-8pt}
\end{figure}
}

Objects of the same category typically share common parts, for example, the seat, backrest, and legs of chairs. These parts form the concise and compact structure of the objects. Extracting this part-level structure from 3D objects is essential in many applications that require part-level prediction, such as shape editing, dynamics simulation, and physical reasoning, including path planning and grasping.

Many supervised approaches~\cite{GRASS-Siggraph-2017, PRNN-ICCV-2017, Im2Struct-CVPR-2018, StructureNet-CVPR-2019, DSGNet-Siggraph-2022} have been proposed to learn a latent structural representation for 3D shapes.
However, these methods require part-level annotations of 3D objects, which are very time-consuming to obtain. 
Unsupervised structural reconstruction methods attempt to learn structural representations of objects by fitting the target shape with a set of geometric primitives, such as cuboids~\cite{VP-CVPR-2017, HA-Siggraph-2019, CA-Siggraph-2021}, planes~\cite{BSPNet-CVPR-2020}, and convexes~\cite{CvxNet-CVPR-2020}.
Although some methods have shown the ability to predict structure-consistent results from unlabeled objects, they tend to generate over-partitioned structures in regions of rich geometric details.
Implicit functions~\cite{occnet-CVPR-2019, IMNET-CVPR-2019, DeepSDF-CVPR-2019} have shown their strong power for modeling geometric details and been applied for structural shape reconstruction~\cite{BAENet-ICCV-2019, RIM-CVPR-2022}. 
However, under the unsupervised learning scheme, these methods extract high-level inter-category common structures from reconstruction error only.  
Without any structural supervision and shape priors, they face significant challenges in approximating detailed shapes in diverse structures.

Taking advantage of the expression power of the implicit functions for geometric details and the generalization ability of geometric primitives to extract high-level structures, we propose a novel deformable primitive field representation that integrates the implicit deformation field and parameterized primitives. 
Considering the shape resemblance between man-made object parts to geometric primitives, we propose a new explicit structure representation based on simple geometric primitives.
In contrast to existing methods that assemble basic geometric primitives to fit the target shape, we consider primitives as structural proxies which can be further deformed for shape refinement.
To support deformation to fit the geometric details of each part, we design a field-based representation for primitives, named parameterized primitive field (PPF), which can be learned end-to-end. 
To improve the expression ability of the primitive fields for detailed geometry, we further concatenate an implicit deformation field to bridge the gap between the abstract structural representation and fine-grained shape details of diverse 3D shapes.

Specifically, we propose DPF-Net, a novel unsupervised structural reconstruction method to approximate target shapes by assembling a set of deformable primitive fields. 
We design a two-stage shape reconstruction pipeline that consists of a Primitive Generation Module (PGM) and a Primitive Deformation Module (PDM) for each part. 
The PGM estimates the explicit orientation, position, and size parameters of parameterized geometric primitives for each part from the encoded shape feature.
These parameterized primitives generated by PGMs can be considered as an intermediate structural representation of the target shape.
Then the PDM predicts dense deformation fields based on the parameterized primitive shape field of each part to recover geometric details.
Each part is assigned a predicted confidence of whether it appears in a specific target shape. 
An occupancy field is finally constructed by assembling all the deformed primitive fields of parts with their confidences.
The whole reconstruction network is trained end-to-end in an unsupervised manner from the shape reconstruction error. 
The strong shape prior of our parameterized primitives allows our DPF-Net to focus on extracting high-level structure at the first module and recovering fine-grained shape details in the second module. 
Combining the explicitly parameterized primitives and implicit deformation fields, we achieve high-quality structural reconstruction for various shapes with consistent structures.

\section{Related Work}
\begin{figure*}[ht]
\centering
\includegraphics[width=1\textwidth]{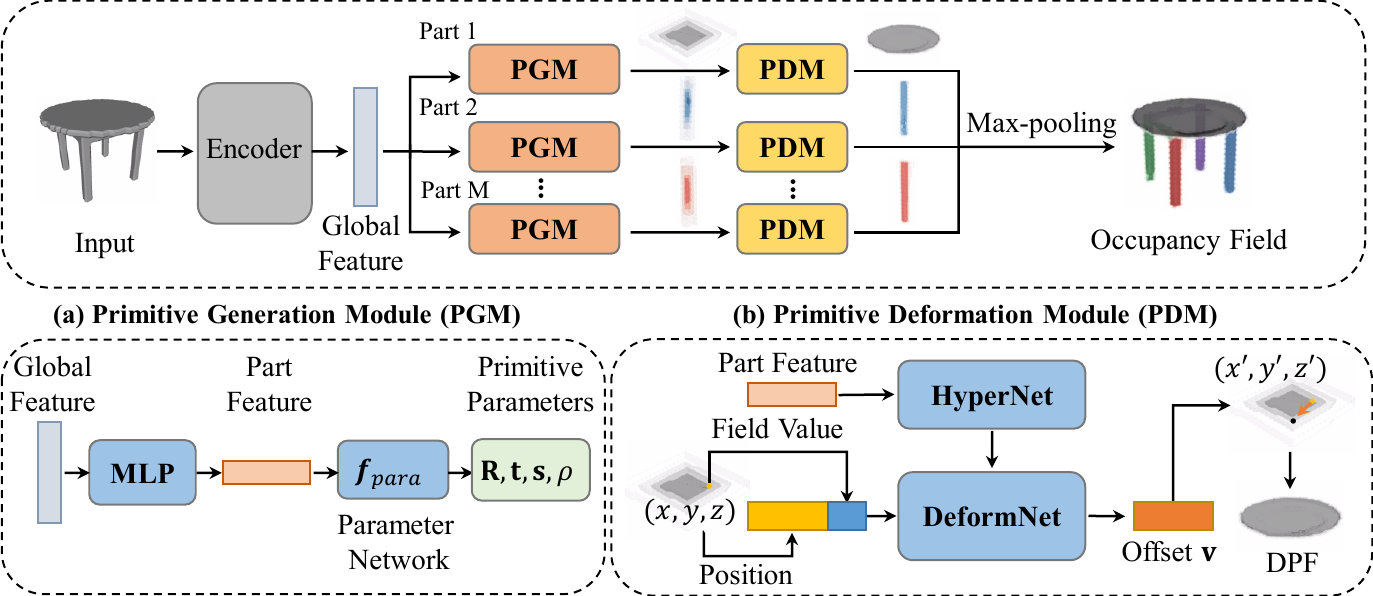}
\caption{Overview of our DPF-Net for unsupervised structural reconstruction. Our DPF-Net mainly consists of two modules: (a) primitive generation module (PGM) and (b) primitive deformation module (PDM). We first project the extracted global feature of the input into a set of part features and generate a parameterized primitive for each part in the primitive generation module. Then for each part, the primitive deformation module predicts point-wise deformation and constructs a deformed primitive field (DPF) to better approximate shape details.
Finally, we compose these DPFs to construct an occupancy field, from which the surface mesh can be extracted by $k$-isosurfacing. }
\label{fig.pipeline} 
 
\end{figure*}

\paragraph{Unsupervised Shape Abstraction.}

Most man-made objects exhibit organized and structured parts, leading to a recent interest in shape abstraction. 
It aims to approximate object shapes by assembling a set of 3D primitives with part-level correspondences across instances. 
Tulsiani \emph{et al.}~\cite{VP-CVPR-2017} proposed the first unsupervised deep-learning-based method that extracts abstracted shapes by predicting the parameters of a set of cuboids with part-level correspondences across instances in an object category. 
Subsequently, researchers try to improve the expression of shape details using the cuboid-based shape reconstruction framework. 
Sun \emph{et al.}~\cite{HA-Siggraph-2019} design an adaptive hierarchical cuboid representation and adaptively select one cuboid in different granularity for each part from multi-level cuboid predictions. 
Yang and Chen~\cite{CA-Siggraph-2021} propose an unsupervised shape abstraction approach for 3D point clouds by jointly learning from point cloud co-segmentation. 
This method achieves more consistent part correspondence and closer shape approximation. 
However, the inherited shape simplicity of the cuboid representation limits its expressiveness of diverse object shapes.

\paragraph{Shape Reconstruction with Convex Elements}
More generalized convex elements~\cite{SQs-CVPR-2019,HSQs-CVPR-2020,CvxNet-CVPR-2020, BSPNet-CVPR-2020} can be used to represent part shapes beyond 3D cuboids. 
BSP-Net~\cite{BSPNet-CVPR-2020} uses planes as the basic shape elements and constructs a BSP-tree structure to generate surface models of 3D objects.
However, it requires a relatively large number of elements to approximate complex shapes. 
To obtain the object parts, they need post-process the partition results by manually grouping the convex elements, which is costly for part-parsing.
Paschalidou \emph{et al.}~\cite{SQs-CVPR-2019, HSQs-CVPR-2020} apply superquadric primitives controlled by two parameters to represent objects, which enhanced the geometric expressiveness for better fitting the curved area of the shape.
However, superquadrics ignore the simplicity of primitive for concise structure, resulting in inconsistent structure extraction.
In contrast, our deformable primitive field uses basic geometry shape as a structural proxy and the implicit deformation function to represent geometry, achieving both structural consistency and high-quality geometric reconstruction.

\paragraph{Structured Implicit Functions}
Implicit functions are more expressive for diverse geometric shapes and object structures than primitive-based and convex elements-based representations and have been explored for unsupervised learning of 3D object structures.
BAE-Net~\cite{BAENet-ICCV-2019} employs a branched implicit function, where each branch represents a part of the object.
This function is restricted as a three-layer MLP to ensure part decomposition ability but limits the capability for expressing complex shapes.
To overcome this, RIM-Net~\cite{RIM-CVPR-2022} proposes a hierarchical branched implicit function with a per-point Gaussian to obtain finer structure extraction.
But RIM-Net struggles to express singular parts due to no explicit structural supervision and shape priors.
Recently, Vasu \emph{et al.}~\cite{HybridSDF-3DV-2022} proposed HybridSDF to combine deep implicit surfaces and geometric primitives.
HybridSDF adopts a disentangled representation that defines the geometric parameters of several parts and has the rest of the object adapt to these specifications.
However, it requires annotating parts as supervision.
Instead, in our DPF-Net, we propose parameterized primitive fields as the intermediate structural proxy and adopt implicit functions to model shape details, generating structurally consistent reconstruction in an unsupervised learning framework.

\paragraph{Shape Deformation} 
Shape deformation~\cite{shapeDeform1-1986, shapeDeform2-2006, shapeDeform3-2010, Atlasnet-CVPR-2018, pixel2mesh-ECCV-2018, shapeflow-NIPS-2020, shapeDeform4-ECCV-2020, neuralcages-CVPR-2020} is a long-term research topic in 3D reconstruction, which model detailed shapes by deforming a template to fit the target shape.
Some of these methods~\cite{Atlasnet-CVPR-2018, pixel2mesh-ECCV-2018, shapeDeform4-ECCV-2020, neuralcages-CVPR-2020} achieve impressive high-quality reconstruction results without taking object structures into account, thus not supporting many downstream applications. Recently, Deng \emph{et al.}~\cite{DIF-CVPR-2021} propose an implicit function-based deformation field that achieves high-quality reconstruction and dense correspondence on the entire shape, but the part-level correspondence is not supported. 
On the other hand, Neural Parts~\cite{NeuralParts-CVPR-2021} achieve local-based deformation by leveraging a group of spheres as part templates and learning homeomorphic mappings between a sphere and the target object. While it mainly focuses on learning more expressive shape primitives, the structural correspondence of the extracted shape segments is ignored. In comparison, our DPF-Net uses primitive filed to extract the structure as the template and leverages implicit deformation field to capture the local geometric details, ensuring high-quality reconstruction and structure extraction.

\section{Our Method}

We follow a typical assembling process that composes 3D objects with a set of parts that share similar geometric structures and layouts across the category. 
We will first introduce our \emph{deformable primitive field} representation for part assembling.
Based on this novel presentation, we design a novel network, named DPF-Net, that reconstructs 3D object shapes by predicting a set of parameterized primitives as parts and then deforming the primitives to fit the target object, as shown in Figure~\ref{fig.pipeline}.
 
\subsection{Representation: Deformable Primitive Field}
\label{sec.method.DPF}

We assemble a 3D object $O$ with $M$ parts $\{P_i\}_{i=1,\ldots, M}$ whose part correspondences across the instances in one category are determined naturally by their orders.
$M$ is a predefined maximum number of the parts in the category.
To impose strong geometric priors on each part, a parameterized geometric primitive, such as a cuboid and a cylinder, can be used to represent the part shape. 
However, this parameterized representation is insufficient to express diverse shape details.
To capture different levels of geometric details of each part, we apply point-wise deformation based on the parameterized primitives. 
Taking a field-based representation, we design a parameterized primitive field (PPF) with strong geometric priors and a deformed primitive field (DPF) to capture fine geometric details. 
 
\subsubsection{Parameterized Primitive Field}

A parameterized primitive field $\mathcal{T}$ encodes a strong shape prior to a part by defining the part shape based on a parameterized geometric primitive, such as a cuboid or a cylinder.
A parameterized geometric primitive $U$ is defined as
\begin{equation} \label{eq:primitive-parameters}
    U = \{\mathbf{R}, \mathbf{t}, \mathbf{s}, \rho\}
\end{equation}
where $\mathbf{R}_{3\times3}$ is a 3D rotation matrix, $\mathbf{t} \in \mathbb{R}^3$ is translation vector, $\mathbf{s} \in \mathbb{R}^3$ is scaling vector, respectively controlling the orientation, position, and size of a predefined unit geometric primitive to compose the target object.
With its primitive parameters $U$, we can transform an arbitrary point $\vb{p}$ in the local coordinate system of this primitive to a point on the object as $\vb{p}=\vb{R}\vb{q}+\vb{t}$.
We define $\rho$ as the confidence of the parameterized primitive $U$ when composing the object since not each of the $M$ parts always appears in a specific object instance. 
Instead of using implicit fields, we define an explicit parameterized primitive field for specific geometric primitives, i.e., cuboid primitives and cylinder primitives, as shown in Figure~\ref{fig.primitiveField}.
This parameterized primitive field measures a structural distance of a 3D point $\vb{p}=(p_x,p_y,p_z)$ to the primitive center under the primitive structure. 

\noindent\textbf{Parameterized Primitive Field of a Cuboid.} For a cuboid primitive with the scaling factor $\vb{s}=(s_x,s_y,s_z)$, its parameterized primitive field $\mathcal{F}^{cub}_{(\vb{s})}$ is defined as
\begin{equation} \label{eq:cuboid-field}
    \ppf^{cub}_{(\vb{s})}(\vb{p}) = \max \left\{ \frac{|p_x|}{s_x},\frac{|p_y|}{s_y},\frac{|p_z|}{s_z} \right\}.
\end{equation}

\noindent\textbf{Parameterized Primitive Field of Cylinder.} For a cylinder primitive with the scaling factor $\vb{s}$, its parameterized primitive field $\mathcal{F}^{cyl}_{(\vb{s})}$ is defined as
\begin{equation}
     \ppf^{cyl}_{(\vb{s})}(\vb{p}) = \max\left\{\frac{\sqrt{p_x^2 + p_y^2}}{s_x}, \frac{|p_z|}{s_z}\right\}.
\end{equation} 
Note that for a cylinder primitive, $s_x=s_y$.

We further apply a normalization function $f_o$ to map the value $d$ of each point in $\ppf$ to a value $o \in (0, 1]$ as 

\begin{equation}\label{eq:normalize-field}
   o=f_o(d) = exp(-\tau d) \in (0, 1],
\end{equation}
where $\tau$ is the temperature parameter to control the decay rate of the field value.
The value $o$ can be considered as the probability of the occupancy of a point in the parameterized primitive field. 

\subsubsection{Deformed Primitive Field}
\label{sec.method.deform}

While the explicit parameterized primitive field $\ppf^{cub}$ or $\ppf^{cyl}$ encodes strong shape priors which can facilitate the part-level structure extraction, it is limited in expressing geometric detail and shape diversity. 
To reconstruct the target part shape with more details from the simple primitives, we further construct an implicit deformation field to bridge the gap between the abstract primitive representation and fine-grained part geometry. 
More specifically, to fit the target object more accurately, we apply a local offset $\vb{v}\in \mathbb{R}^{3}$ for each point $\vb{q}$ in the parameterized primitive space. 
We predict the point-wise offset field of each part from the point positions and the extracted part feature. 
By applying the point-wise deformation, a deformed primitive field $\ppf^{def}$ can be constructed as $\ppf_{(\vb{s})}(\vb{q}+\vb{v})$ under the structural priors of each primitive type, as shown Figure~\ref{fig.primitiveField} (b).

\subsubsection{Assembling Primitive Fields}
\begin{figure}
  \centering
  \includegraphics[width=0.45\textwidth]{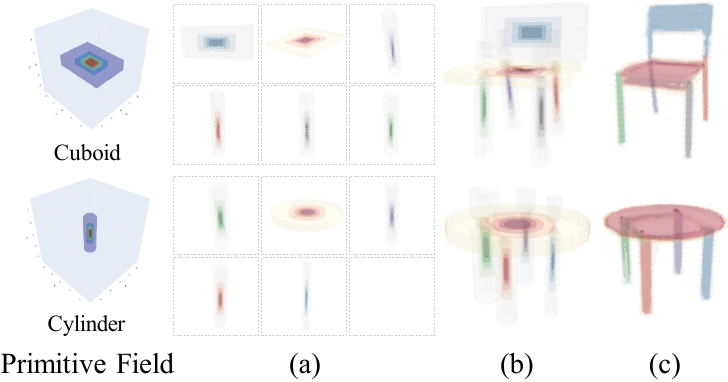}
  \caption{Illustration of primitive fields. (a) Each object part is represented by a parameterized primitive field of a cuboid primitive (top) or a cylinder primitive (bottom) with different parameters. (b) The parameterized primitive is further deformed via a deformation field to better approximate the shape details of each part. (c) An occupancy field is obtained for the entire object by assembling the deformed primitive fields of all the parts.
} 
  \label{fig.primitiveField} 
\end{figure}

Then we assemble a 3D object by the $M$ parts by composing the $M$ normalized deformed primitive fields together, as Figure~\ref{fig.primitiveField} shows.
Since we do not have the deterministic assignment of the $M$ primitives to each point on the object, we introduce a confidence $\rho \in [0,1]$ for each part, as defined in Eq.~\ref{eq:primitive-parameters}. 
For each point $\mathbf{q}_j$ on the target object, we transform it to a point $\vb{p}_{ji}=\vb{R}^{-1}_i(\vb{q}_j-\vb{t}_i)$ in the local coordinate system of each parameterized primitive with parameters $U_i$ and compute its normalized field value after further deforming it by the offset $\vb{v}_{ji}$ as
\begin{equation} \label{eq:object-primitive-field}
\mathcal{F}^{obj}_i(\vb{q}_j)= \rho_i \cdot f_o \big( \ppf_{(\vb{s_i})}(\vb{p}_{ji}+\vb{v}_{ji}) \big),
\end{equation}
where $\rho_i$ is the confidence of the primitive $U_i$.
 
We apply max-pooling of the $M$ deformed primitive fields $\mathcal{F}^{obj}_i$ for each object point, thus getting the final assembled occupancy field $\ppf^{obj}$. 
The object surface mesh can be generated by taking the $k-$isosurface of the occupancy field using Marching Cubes~\cite{MC-IJCAI-1977}.

\dlt{
\noindent \textbf{Discussion.}
According to the above, by feeding a point cloud or volume to the encoder, we can get a group of primitive fields generated from part features, of which each represents a single part of an object.
Based on the position $\mathbf{t}$ and direction $\mathbf{r}$ of each parameterized primitive field, we can pose those primitive fields in the same spatial space together, which can be regarded as the structure of the input.
Noted that this design injects inductive biases about the ability to understand objects from the structural perspective.
Since artifacts always reflect similar design commonalities in their structure, our method can learn the semantic consistent structure from a class of objects from a structural perspective during training.
}

\subsection{Structural 3D Reconstruction}
\label{sec:reconstruction}

Based on our novel deformable primitive field representation, we now explain how to reconstruct 3D objects from a point cloud or a 3D volume. 
We first extract a global feature from the input volume via a 3D neural network similar to BSP-Net~\cite{BSPNet-CVPR-2020}. 
Then we project the global feature into $M$ part features $\{\mathbf{F}_i \in \mathbb{R}^n\}_{i=1,\ldots,M}$ using $M$ multi-layer perceptrons (MLP) for each primitive. 
Next, we predict the primitive parameters $U_i$ for each geometric primitive from the part feature $\mathbf{F}_i$ using weight-sharing MLP $f_{para}$.

In the primitive generation module, we estimate the parameters for each geometric primitive and produce a PPF of each part as a coarse approximation of the target shape.
Then in the primitive deformation module, we further predict the offset $\vb{v}_{ji}$ of each 3D point $\vb{q}_j$ sampled in the object space to better fit the target shape. 
For each part branch, taking each point $\vb{q}_{j}$ and the parameterized primitive field $\ppf_i$ as input, we employ a deformation network, denoted as DeformNet $D_i$, to predict the offset $\vb{v}_{ji}$. 
Instead of directly training an MLP-based DeformNet for each part, we employ a hypernetwork scheme that has been studied for structure-aware 3D representation~\cite{SRN-NIPS-2019, DIF-CVPR-2021}. 
A hyper-net $\Psi$ is first applied to predict the network weights $\vb{\omega}_i$ from the part feature $\mathbf{F}$. 
Then the deformer $D_i$ uses these weights $\vb{\omega}_i$ to generate the point-wise offset $\mathbf{v}_{ji}$ for a 3D point $\vb{q}_j$.
Moreover, DIF-Net~\cite{DIF-CVPR-2021} introduces an additional correction scalar $c$ to correct their template fields for diverse structures.
We inherit this correction scalar in our deformation network by adding the point-wise correction value $c_{ji} \in [-1, 1]$ to our deformed primitive field of each part $\ppf^{obj}_i$. 
\begin{equation}
    \Psi: \mathbf{F}_i  \rightarrow \vb{\omega}_i,
\end{equation}
\begin{equation}
    D^{(\vb{\omega_i})}: (\mathbf{q}_j, o_{ji}) \rightarrow (\mathbf{v}_{ji}, c_{ji}),
\end{equation}
where $\mathbf{q}_j$ is a point in 3D space and $o_{ji}$ is its corresponding value in the parameterized primitive field of the $i$th part.

\begin{table*}
 \setlength\tabcolsep{8.0pt}
  \centering
    \caption{Quantitative comparison of unsupervised structural reconstruction performance on three object categories in ShapeNet. }
    \label{tab.recon}
    \begin{tabular}{l||cccc||cccc}
    \toprule
      & \multicolumn{4}{c||}{Chamfer Distance (CD) ($\times 0.001$) $\downarrow$ } & \multicolumn{4}{c}{Intersection over Union (IoU) ($\%$) $\uparrow$} \\
    \cmidrule{2-9} 
    Method & table & chair & airplane & Mean  & table & chair & airplane & Mean \\
    \midrule
    \midrule
    VP~\cite{VP-CVPR-2017}    & 1.1561 & 0.7989 & 0.4587 & 0.8046 & 68.61 & 73.82 & 72.68 & 71.70  \\
    HA~\cite{HA-Siggraph-2019}    & 1.0034 & 0.7199 & 0.2802 & 0.6678 & 72.25 & 77.56 & \emph{77.35} & 75.72 \\
    CA~\cite{CA-Siggraph-2021}    & 0.9952 & 0.6682 & 0.3609 & 0.6748 & 68.46 & 72.32 & 69.25 & 70.01 \\
    BAE-Net~\cite{BAENet-ICCV-2019}   & 1.2775 & 0.6945 & 0.4276 & 0.7999 & 62.57 & 72.29 & 69.34 & 68.07 \\
    RIM-Net~\cite{RIM-CVPR-2022}  & 0.7463 & 0.4125 & 0.2228 & 0.4605 & 75.85 & 79.61 & 74.53  & 76.66 \\
    
    BSP-Net~\cite{BSPNet-CVPR-2020} & \emph{0.5140} & \emph{0.3420} & \textbf{0.1350} & \textbf{0.3303}  & \textbf{79.23} & \emph{80.20} &74.21 & \emph{77.88} \\
     DPF-Net (Ours)  & \textbf{0.4756
} & \textbf{0.3633} & \emph{0.1924} & \emph{0.3438} & \emph{76.11} & \textbf{80.91} & \textbf{79.08} & \textbf{78.70}  \\
 
    \bottomrule
    \end{tabular} 
\end{table*}

\begin{figure*}
  \centering
\includegraphics[width=1\textwidth]{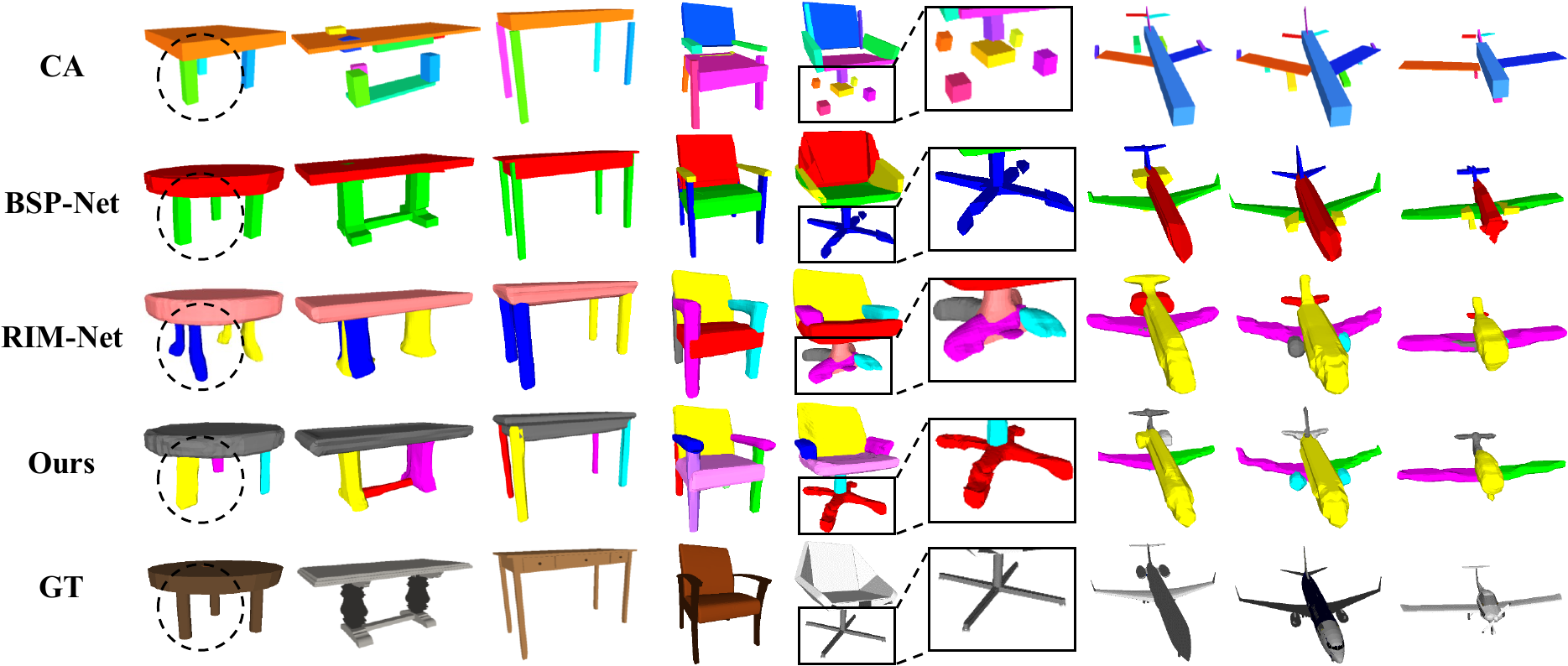}
  \caption{Qualitative comparison of unsupervised structural reconstruction methods.}
  \label{fig.reconstruction}
\end{figure*}

\subsection{Network Training}
\label{sec:network-training}

Our model is trained by optimizing the reconstruction error of the target shape using our deformable primitive representation, without any part-level annotations. 
The overall loss function $L$ is composed of four loss items:
\begin{equation}
L = \lambda_1 L_{recon}+\lambda_2L_{deform}+ \lambda_3 L_{comp}+\lambda_4 L_{align},
\end{equation}
where the weights $\{\lambda_i\}$ balance the four loss terms.

\noindent\textbf{Reconstruction Loss.}
The reconstruction loss measures the distance from the predicted shape to the target shape. 
Same as~\cite{BAENet-ICCV-2019}, we compute the distance between the predicted occupancy field $\ppf^{obj}$ and the ground-truth occupancy field $\mathcal{O}^{gt}$ of the target shape.
$\mathcal{O}^{gt}(\mathbf{q}) \in \left\{0, 1\right\}$ at location $\mathbf{q}$.
We sample $N_{\Omega}$ points in the 3D space surrounding the input shape and compute the root mean square error of the sampled point set $\Omega$ as the reconstruction loss:
\begin{equation}
    L_{recon} = \frac{1}{N_{\Omega}}\sum_{\mathbf{q} \in \Omega} \vert\vert \ppf^{obj}(\mathbf{q})-\mathcal{O}^{gt}(\mathbf{q}) \vert\vert_2.
\end{equation}

\noindent\textbf{Deformation Regularization Loss.}
Our DeformNet predicts the deformation offset $\vb{v}_{ji}$ for each point to capture the detailed geometry of each part.
However, to enforce the shape constraint of the primitives and restrict the deformed shape to follow shape regularity of the predicted primitives, we design a deformation regularization loss:
\begin{equation}
    L_{deform} = \frac{1}{N_{\Omega}}\sum_{j}^{N_{\Omega}} \sum_i^M \vert\vert \mathbf{v}_{ji} \vert\vert_2.
\end{equation}
 
\noindent\textbf{Compactness Loss.} 
Objects of different categories might have various numbers of parts. In our DPF-Net, we set a maximum part number $M$ for various objects. However, we want to make the learned object structure as compact as possible to prevent over-partition. We employ the compactness loss proposed in CA~\cite{CA-Siggraph-2021}. 
We sample $N_p$ points $P=\{\vb{q}_{j=1,\ldots,N_p}\}$ on the surface mesh of the target shape and then compute the final occupancy probability $\ppf_i^{obj}(\vb{q}_j)$ of each point $\vb{q}_j$ by each part. 
$L_{0.5}$ normalization term for assignment matrix $\vb{W}$ is computed as the compactness loss to constrain its sparsity.
\begin{equation}
    L_{comp} = \left(\sum_{i=1}^M\sqrt{\frac{1}{N_P}\sum_{\vb{q}_j\in P} \ppf_i^{obj}(\vb{q}_j) + \epsilon}\right)^2,
\end{equation}
where $\epsilon$ is a small constant to prevent gradient explosion. 
 
\noindent\textbf{Axis-Aligned Alignment Loss.}
Since many parts of man-made objects are typically axis-aligned, we employ the axis-aligned alignment loss proposed in HA~\cite{HA-Siggraph-2019}.
For the predicted rotation matrix $\mathbf{R}_i$ of each primitive, its corresponding quaternion $\mathbf{r}_i$ is enforced to be close to the identity transformation $\mathbf{I} = (1,0,0,0)^\top$ by
\begin{equation} 
    L_{align} = \frac{1}{M}\sum_i^M \vert\vert \mathbf{r}_i - \mathbf{I} \vert\vert_2.
\end{equation}

\section{Experiments and Results}
We conducted extensive experiments to demonstrate the effectiveness of our proposed DPF-Net in reconstructing structural models, including comparison with state-of-the-art unsupervised shape abstraction methods and ablation studies. We evaluate our DPF-Net and related methods on the ShapeNet part dataset~\cite{shapenet-2015} using three categories, including airplane $(2,690)$, chair $(3,758)$, and table $(5,271)$. 

\subsection{Experimental Settings}

\noindent\textbf{Training Details.}
For model training, we set the temperature parameter $\tau = 4$ used in Eq.~\ref{eq:normalize-field} for field value normalization. 
The four loss weights are $\lambda_1 = 1$, $\lambda_2 = 0.1$, $\lambda_3 = 0.0001$, and $\lambda_4 = 0.0001$ respectively.
We train our DPF-Net using Adam~\cite{Adam-2014} optimizer with $lr=0.0001$, $\beta_1 = 0.5$, $\beta_2 = 0.9$.
For each object category, we train a model progressively with the input volumes of $32^3$ and $64^3$ resolutions. 
The number of sampled points $N_{\Omega}=8,192$ for $32^3$ resolution and $N_{\Omega}=32,768$ for $64^3$ resolution.
For the compactness loss, $N_P=1024$. 
For each resolution, the model is trained for $1000$ epochs with a batch size of $128$.
The total training time for each category is about $6$ hours with two NVIDIA A100 GPUs.
The surface meshes are constructed as 0.6-isosurface of our final occupancy field at $64^3$ resolution by Marching Cubes~\cite{MC-IJCAI-1977}.


\noindent\textbf{Metrics.}
We conduct quantitative and qualitative evaluations on two tasks, i.e., structural shape reconstruction and shape segmentation. 
For reconstruction, Chamfer Distance (CD)~\cite{MC-IJCAI-1977} and Intersection over Union (IoU) are used to evaluate the reconstruction precision.
Specifically, CD is calculated using $4,096$ uniformly sampled points from the surface of both the ground-truth shape and reconstructed mesh, and the IoU is calculated on the volumes of $32^3$ resolution.
For part segmentation, same as~\cite{BAENet-ICCV-2019}, we adopt the mean per-label Intersection over Union (m-IoU) as the metric to measure the quality and consistency of the recovered parts.
 
\begin{table}[tbp]
\setlength\tabcolsep{6.0pt}
  \centering
     \caption{Quantitative comparison of segmentation m-IoU ($\uparrow$).} 
    \begin{tabular}{l||c|c|c}
    \toprule
    Shape (\#parts) & table (2) & chair (4) & airplane (4) \\
    \midrule
    \begin{tabular}[c]{@{}c@{}}Segmented\\ parts\end{tabular} & \begin{tabular}[c]{@{}c@{}}top,\\ support\end{tabular} & \begin{tabular}[c]{@{}c@{}}back, seat,\\ leg, arm\end{tabular} & \begin{tabular}[c]{@{}c@{}}body, tail,\\ wing, engine\end{tabular}  \\
    \midrule
    VP~\cite{VP-CVPR-2017}     & 62.1  & 64.7 & 37.6   \\
    HA~\cite{HA-Siggraph-2019}   & 67.4 & 80.4 & 55.6  \\
    CA~\cite{CA-Siggraph-2021}   & 89.2 & 82.0 & 64.2  \\
    BAE-Net~\cite{BAENet-ICCV-2019}  & 87.0 & 65.5 & 61.1  \\
    BSP-Net~\cite{BSPNet-CVPR-2020} & 90.3 & 80.9 & \textbf{74.2} \\
    RIM-Net~\cite{RIM-CVPR-2022} & 91.2 & 81.5 & 67.8 \\
    DPF-Net (Ours)                       & \textbf{91.3} & \textbf{84.3} & 66.0  \\
    \bottomrule
    \end{tabular}
    \vspace{-12pt}
    \label{tab.seg}
\end{table}%

\subsection{Structural Shape Reconstruction}
We evaluate our model on the shape reconstruction task compared with unsupervised shape abstraction methods, including cuboid-based methods~\cite{VP-CVPR-2017, HA-Siggraph-2019, CA-Siggraph-2021} and implicit function methods~\cite{BAENet-ICCV-2019,BSPNet-CVPR-2020, RIM-CVPR-2022}.
We train our model with the predefined maximum number of parts $M=16$ for chairs, $M=8$ for tables and airplanes empirically.
We test two types of primitives and mainly use the results obtained using cuboid primitives for comparison with other methods.

\noindent \textbf{Quantitative analysis.}
Table~\ref{tab.recon} shows the quantitative comparison results of shape reconstruction.
The cuboid-based methods VP~\cite{VP-CVPR-2017}, HA~\cite{HA-Siggraph-2019}, and CA~\cite{CA-Siggraph-2021} have trouble capturing diverse geometric details due to the limited expression ability of cuboids.
Regarding the implicit function-based methods, while BAE-Net~\cite{BAENet-ICCV-2019} does not perform well due to its simple three-layer MLP architecture, RIM-Net~\cite{RIM-CVPR-2022} improves the reconstruction performance by its hierarchical presentation. 
BSP-Net~\cite{BSPNet-CVPR-2020} achieves better reconstruction quality by generating hundreds of convexes from thousands of planes to fit the target shapes with fine-grained details. 
In comparison, our DPF-Net achieves comparable CD and higher IoU with BSP-Net with much fewer primitives, conveying more accurate and explicit structural correspondences between semantic parts.
 

\noindent \textbf{Qualitative analysis.}
We show reconstruction results for the three categories in Figure~\ref{fig.reconstruction}. 
The cuboid-based shape abstraction method CA well recovers the structure of objects that are mainly composed of parts in cuboid shape.
But for more complex part shapes, CA tends to generate a lot of cuboids to fit the target shape, thus leading to over-partition, such as the second table and the second chair example.
RIM-Net can approximate various shapes that follow Gaussian point distributions but fails to reconstruct particular parts. 
For the table samples, most of the training examples have symmetric legs, so RIM-Net tends to generate symmetric shapes even for the three legs in the first example. 
CA~\cite{CA-Siggraph-2021} and RIM-Net~\cite{RIM-CVPR-2022} both have difficulty capturing shape details for particular structures, such as the chair legs. CA suffers over-partition in regions of rich shape details. RIM-Net fails to model the shape of singular parts, e.g., the three legs of the first table and the beam between the legs. In comparison, our reconstruction results show consistent structural partition while recovering unique part shapes, such as the beam of the second table example and the wheels of the second chair example.
BSP-Net~\cite{BSPNet-CVPR-2020} better reconstructs the target shape by assembling massive primitives but leads to corrugated surfaces, such as the back of the second chair and the airplane head.
With fewer primitives, our DPF-Net generates high-quality 3D reconstruction with smooth surface meshes and consistent part structures.

\begin{figure}
  \centering
\includegraphics[width=\linewidth]{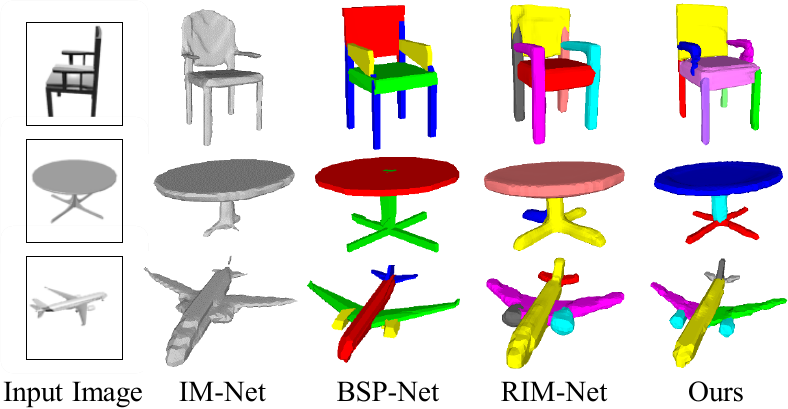}
\caption{Qualitative comparison on SVR task. Our DPF-Net better reconstructs the target shapes (chair arms, table legs, airplane engines) with a more reasonable part partition. }
\label{fig.svr}
\vspace{-1pt}
\end{figure}



\subsection{3D Shape Segmentation}
\label{sec.exp.seg}
We compare the unsupervised shape segmentation performance with cuboid-based methods~\cite{VP-CVPR-2017, HA-Siggraph-2019, CA-Siggraph-2021} and implicit function methods~\cite{BAENet-ICCV-2019, BSPNet-CVPR-2020, RIM-CVPR-2022} to demonstrate the structural consistency of our reconstructed shapes.
Following RIM-Net~\cite{RIM-CVPR-2022}, the part annotations in the ShapeNet dataset are merged to fewer groups. For example, ``leg'' and ``support'' of tables are merged into ``support''.
While there are no semantic part annotations for the unsupervised segmentation task, we assign each primitive with semantics following the common settings~\cite{CA-Siggraph-2021,RIM-CVPR-2022} and compute the per-label m-IoU.
As shown in Table~\ref{tab.seg}, our method outperforms the others on the chair and table categories which exhibit more diverse structures, demonstrating the superiority of our DPF-Net in recovering consistent structures across various shapes. 
For airplanes, our method achieves comparable performance to RIM-Net but slightly underperforms BSP-Net. 
This is mainly because BSP-Net groups many primitives as one semantic part, while mixing the shape correspondences. 
In contrast, our method uses much less primitives and the structural corespondences are naturally encoded by the primitive deformation.

\subsection{Single-View Reconstruction}

We also apply our DPF-Net for single-view image reconstruction to further demonstrate the scalability of our structural shape representation.
We use chair, table and airplane data from ShapeNet~\cite{shapenet-2015} with rendered views by 3D-R2N2\cite{3D-R2N2-ECCV-2016}.
The train/test splitting is the same as ~\cite{BSPNet-CVPR-2020}.
Following RIM-Net~\cite{RIM-CVPR-2022}, we first pretrain a 3D auto-encoder for each category.
A 2D image encoder is trained to extract image features by minimizing the mean square error between the extracted image feature and the features extracted by the pretrained 3D auto-encoder. Table~\ref{tab.svr} and Figure~\ref{fig.svr} show the quantitative and qualitative comparisons respectively.
Our DPF-Net ourperforms IM-Net~\cite{IMNET-CVPR-2019}, BSP-Net~\cite{BSPNet-CVPR-2020}, and RIM-Net~\cite{RIM-CVPR-2022} averagely on the three categories. All methods perform similar results in the chair category due to the occlusion problem of chair images. Our DPF-Net obtains 3D shapes closer to the input image with better part partition.

\begin{table}[t!]
  \setlength\tabcolsep{2.0pt}
  \centering
  \caption{Single view reconstruction performance in CD ($\times 0.001$).}%
  \label{tab.svr}%
    \begin{tabular}{l||ccc|c}
\toprule
Methods & table & chair & airplane & mean\\
    \midrule
    \midrule
    IM-Net~\cite{IMNET-CVPR-2019}  & \textbf{0.8762} & 0.8799  & 0.4041 & 0.7201\\
    BSP-Net~\cite{BSPNet-CVPR-2020}  & 0.9807 & 0.7472  & 0.4716 &  0.7332 \\
    RIM-Net~\cite{RIM-CVPR-2022}  & 1.1223 & \textbf{0.7446}  & 0.3993 & 0.7554  \\
    Ours  & \emph{0.9029} & \emph{0.7466}  & \textbf{0.3840} & \textbf{0.6847}  \\
    \bottomrule
    \end{tabular}%
\end{table}%

\subsection{Ablation Study}

\begin{figure}
  \centering
  \includegraphics[width=0.45\textwidth]{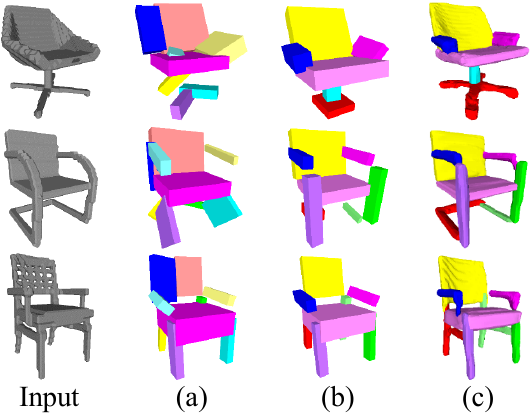}
  \caption{Reconstruction results of different model variants. (a) Model without PDM. (b) Results of intermediate parameterized primitives in our full model. (c) Results of our full DPF-Net.}
  \label{fig.ablation}
  \vspace{-8pt}
\end{figure}

\noindent
\textbf{Deformed Primitive Field.} 
To explore the effect of the proposed deformed primitive field representation, we compare the reconstruction results generated by our full DPF-Net and a variant with the parameterized primitive field only.
In the model ``w/o PDM", we directly regress the primitive parameters to approximate the target shape without using the deformation module. 
In the model `PPM', We approximate the target shapes using only the intermediate parameterized primitives in our full model. 
As Table~\ref{tab.ablation.deform} shows, the intermediate PPM show more consistent part structure (higher m-IoU) but lower reconstruction quality (higher CD). Our full model performs best on both reconstruction and structural partition, by decomposing the challenging structural reconstruction task for diverse shapes into two steps.
With strong shape priors, the parameterized primitive generation module focuses on extracting consistent structures among instances.
Based on the superior structural partition, the primitive deformation module focuses on reconstructing diverse shape details of each individual part.
For the parts with rich geometry detail, without PDM, the network with only parameterized primitives tends to sacrifice the structural consistency for better reconstruction quality using limited primitives, as Figure~\ref{fig.ablation} (a) shows. 
In comparison, while letting the deformation module approximate the shape details, the generated primitives in the first step of our full DPF-Net capture better global part structures (Figure~\ref{fig.ablation} (b)). 
Integrating the deformation module, our DPF-Net finally reconstructs the target shape accurately for objects in diverse structures, as Figure~\ref{fig.ablation} (c) shows.

\begin{table}
 \setlength\tabcolsep{2.0pt}
  \centering
  \caption{Ablation study of three variants of DPF-Net. }
    \label{tab.ablation.deform}
    \begin{tabular}{l|cc|cc|cc}
    \toprule
      & \multicolumn{2}{c|}{table}  & \multicolumn{2}{c|}{chair}  & \multicolumn{2}{c}{airplane} \\
    \cmidrule{2-7} 
    Method & CD & m-IoU & CD  & m-IoU & CD & m-IoU \\
    \midrule
    \midrule
    w/o PDM & 0.7356 &  85.6 & 0.5286 & 79.4 & 0.2412  & 56.3 \\
    PPM  & 0.8265 &  89.8 & 0.8127 & 82.4 & 0.2783   & 58.4  \\
    Full Model & \textbf{0.4756} & \textbf{91.3} & \textbf{0.3633} & \textbf{84.3} & \textbf{0.1941
}  & \textbf{66.0}   \\
 
    \bottomrule
    \end{tabular}%
\end{table}

\begin{figure}
  \centering
  \includegraphics[width=0.45\textwidth]{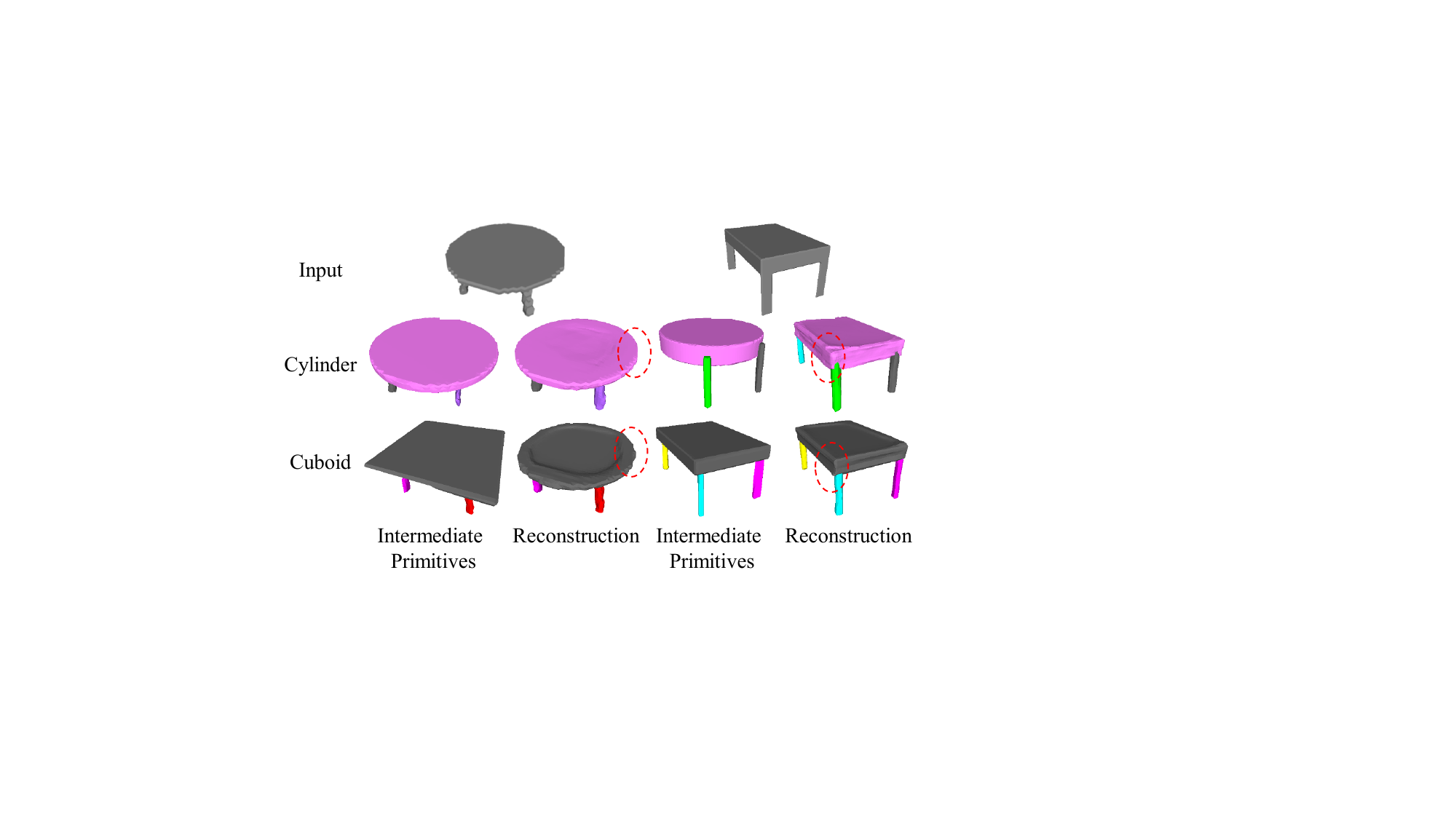}
  \caption{Ablation experiment of different primitive types.}
  \label{fig.ablation.primitivetype}
\end{figure}

\noindent \textbf{Primitive Type.} Our DPF-Net supports different pre-geometric primitives to impose different shape priors. We test our method with cuboid and cylinder primitives separately and report the quantitative results in Table~\ref{tab.ablation.primitivetype}.
Figure~\ref{fig.ablation.primitivetype} shows the reconstruction results and the intermediate primitives using cuboid and cylinder primitives for two table examples.
While the reconstruction results using different primitive types show superiority on specific part shapes that follow the shape priors, for example, cylinders are better for the first table and cuboids are better for the second table, the global structure can be consistently extracted with either type of primitives. 
The final reconstruction results using different primitive types are very close, verifying the robustness and flexibility of our DPF-Net. 
\begin{table}
 \setlength\tabcolsep{2.0pt}
  \centering
  \caption{Quantitative results using different primitives.}
    \label{tab.ablation.primitivetype}
        \begin{tabular}{lcc}
        \midrule
         Primitive Type   & CD   & m-IoU  \\
        \midrule
        Cuboid    & 0.4756 & \textbf{91.3} \\
        Cylinder    & \textbf{0.4251} & 90.4 \\
        \midrule
        \end{tabular} 
\end{table}

\section{Conclusion}
We introduce DPF-Net, an unsupervised structural shape reconstruction method to approximate detailed shapes in diverse structures. The proposed deformable primitive field representation imposes strong shape priors via parameterized geometric primitives to enforce global structure consistency and restores local shape details via a deformation field from geometric primitives. 
By decomposing the structural reconstruction task into a primitive generation module and a primitive deformation module, our DPF-Net can effectively prompt its structure extraction ability of diverse structures and expression capability for shape details. 
The reconstruction results of our DPF-Net on the ShapeNet show high visual quality and consistent structures.

Our DPF-Net is flexible and robust to support different types of geometric primitives. We achieve comparable results using both cuboid and cylinder primitives. In the future, we would like to extend our DPF-Net to support more primitive types. Integrating multiple geometric primitives in the same network is a possible direction to better express the local part shapes in various structures. One limitation of our current DPF-Net is that our reconstructed surfaces are not smooth enough. We think adding a penalty on large deformation gradients, like DIF-Net~\cite{DIF-CVPR-2021}, will be helpful to improve surface smoothness. 

\paragraph{Acknowledgement} 
This work was supported by the National Natural Science Foundation of China under Grant 62076230 and in part by the Fundamental Research Funds for the Central Universities under Grant WK3490000008.  

{\small
\bibliographystyle{ieee_fullname}
\bibliography{chapters/reference}
}

\end{document}